\pdfoutput=1
\documentclass[10pt]{article}
\usepackage{graphicx}

\newcommand{\ra}{\rangle}
\newcommand{\la}{\langle}
\newcommand{\ttt}{\texttt}

\begin{document}
 \title{Using RDF to Model the Structure and Process of Systems\footnote{Rodriguez, M.A., Watkins, J.H., Bollen, J., Gershenson, C., ``Using RDF to Model the Structure and Process of Systems", International Conference on Complex Systems, Boston, Massachusetts, October 2007.}}
 \author{Marko A. Rodriguez \\
Jennifer H. Watkins \\
Johan Bollen \\
Los Alamos National Laboratory\\\{marko,jhw,jbollen\}@lanl.gov \\\\
Carlos Gershenson \\
New England Complex Systems Institute\\carlos@necsi.org
}
\maketitle{}
\abstract{Many systems can be described in terms of networks of discrete elements and their various relationships to one another. A semantic network, or multi-relational network, is a directed labeled graph consisting of a heterogeneous set of entities connected by a heterogeneous set of relationships. Semantic networks serve as a promising general-purpose modeling substrate for complex systems. Various standardized formats and tools are now available to support practical, large-scale semantic network models. First, the Resource Description Framework (RDF) offers a standardized semantic network data model that can be further formalized by ontology modeling languages such as RDF Schema (RDFS) and the Web Ontology Language (OWL). Second, the recent introduction of highly performant triple-stores (i.e.~semantic network databases) allows semantic network models on the order of $10^9$ edges to be efficiently stored and manipulated. RDF and its related technologies are currently used extensively in the domains of computer science, digital library science, and the biological sciences. This article will provide an introduction to RDF/RDFS/OWL and an examination of its suitability to model discrete element complex systems.}

                           
\section{Introduction}

The figurehead of the Semantic Web initiative, Tim Berners-Lee, describes the Semantic Web as 
\begin{quote}
... an extension of the current web in which information is given well-defined meaning, better enabling computers and people to work in cooperation \cite{lee:semantic2001}.
\end{quote}
However, Berners-Lee's definition assumes an application space that is specific to the ``web" and to the interaction between humans and machines. More generally, the Semantic Web is actually a conglomeration of standards and technologies that can be used in various disparate application spaces. The Semantic Web is simply a highly-distributed, standardized semantic network (i.e.~directed labeled network) data model and a set of tools to operate on that data model. With respect to the purpose of this article, the Semantic Web and its associated technologies can be leveraged to model and manipulate any system that can be represented as a heterogeneous set of discrete elements connected to one another by a set of heterogeneous relationships whether those elements are web pages, automata, cells, people, cities, etc. This article will introduce complexity science researchers to a collection of standards designed for modeling the heterogeneous relationships that compose systems and technologies that support large-scale data sets on the order to $10^9$ edges.

This article has the following outline. Section \ref{sec:rdf} presents a review of the Resource Description Framework (RDF). RDF is the standardized data model for representing a semantic network and is the foundational technology of the Semantic Web. Section \ref{sec:onto} presents a review of both RDF Schema (RDFS) and the Web Ontology Language (OWL). RDFS and OWL are languages for abstractly defining the topological features of an RDF network and are analogous, in some ways, to the database schemas of relational databases (e.g.~MySQL and Oracle). Section \ref{sec:store} presents a review of triple-store technology and its similarities and differences with the relational database. Finally, Section \ref{sec:comp} presents the semantic network programming language Neno and the RDF virtual machine Fhat.

\section{The Resource Description Framework\label{sec:rdf}}

The Resource Description Framework (RDF) is a standardized data model for representing a semantic network \cite{rdfspec:manola2004}. RDF is not a syntax (i.e.~data format). There exist various RDF syntaxes and depending on the application space one syntax may be more appropriate than another. An RDF-based semantic network is called an RDF network. An RDF network differs from the directed network of common knowledge because the edges in the network are qualified. For instance, in a directed network, an edge is represented as an ordered pair $(i,j)$. This relationship states that $i$ is related to $j$ by some unspecified type of relationship. Because edges are not qualified, all edges have a homogenous meaning in a directed network (e.g.~a coauthorship network, a friendship network, a transportation network). On the other hand, in an RDF network, edges are qualified such that a relationship is represented by an ordered triple $\la i, \omega, j \ra$. A triple can be interpreted as a statement composed of a subject, a predicate, and an object. The subject $i$ is related to the object $j$ by the predicate $\omega$. For instance, a scholarly network can be represented as an RDF network where an article cites an article, an author collaborates with an author, and an author is affiliated with an institution. Because edges are qualified, a heterogeneous set of elements can interact in multiple different ways within the same RDF network representation. It is the labeled edge that makes the Semantic Web and the semantic network, in general, an appropriate data model for systems that require this level of description.

In an RDF network, elements (i.e.~vertices, nodes) are called resources and resources are identified by Uniform Resource Identifiers (URI) \cite{uri:berners2005}. The purpose of the URI is to provide a standardized, globally-unique naming convention for identifying any type of resource, where a ``resource" can be anything (e.g.~physical, virtual, conceptual, etc.). The URI allows every vertex and edge label in a semantic network to be uniquely identified such that RDF networks from disparate organizations can be unioned to form larger, and perhaps more complete, models. The Semantic Web can span institutional boundaries to support a world-scale model. The generic syntax for a URI is
\begin{verbatim}
	<scheme name> : <hierarchical part> [ # <fragment> ]
\end{verbatim}
Examples of entities that can be denoted by a URI include:
\begin{itemize}\addtolength{\itemsep}{-1.2\baselineskip}
	\item a physical object (e.g.~\ttt{http://www.lanl.gov/people\#marko}) \\
	\item a physical component (e.g.~\ttt{http://www.lanl.gov/people\#markos\_arm}) \\
	\item a virtual object (e.g.~\ttt{http://www.lanl.gov/index.html}) \\
	\item an abstract class (e.g.~\ttt{http://www.lanl.gov/people\#Human}).
\end{itemize}

Even though each of the URIs presented above have an \ttt{http} schema name, only one is a Uniform Resource Locator (URL) \cite{uri:2001} of popular knowledge: namely, \ttt{http://www.lanl.gov/index.html}. The URL is a subclass of the URI. The URL is an address to a particular harvestable resource. While URIs can point to harvestable resources, in general, it is best to think of the URI as an address (i.e.~pointer) to a particular concept. With respects to the previously presented URIs, Marko, his arm, and the class of humans are all concepts that are uniquely identified by some prescribed globally-unique URI.

Along with URI resources, RDF supports the concept of a literal. Example literals include the integer $1$, the string ``marko", the float (or double) $1.034$, the date 2007-11-30, etc. Refer to the XML Schema and Datatypes (XSD) specification for the complete classification of literals \cite{xsd:biron2004}.

If $U$ is the set of all URIs and $L$ is the set of all literals, then an RDF network (or the Semantic Web in general) can be formally defined as\footnote{Note that there also exists the concept of a blank node (i.e.~anonymous node). Blank nodes are important for creating $n$-ary relationships in RDF networks. Please refer to the official RDF specification for more information on the role of blank nodes.}
\begin{equation}
G \subseteq \la U \times U \times (U \cup L) \ra.
\end{equation}

\begin{figure}[h!]
	\begin{center}
		\includegraphics[width=0.55\textwidth]{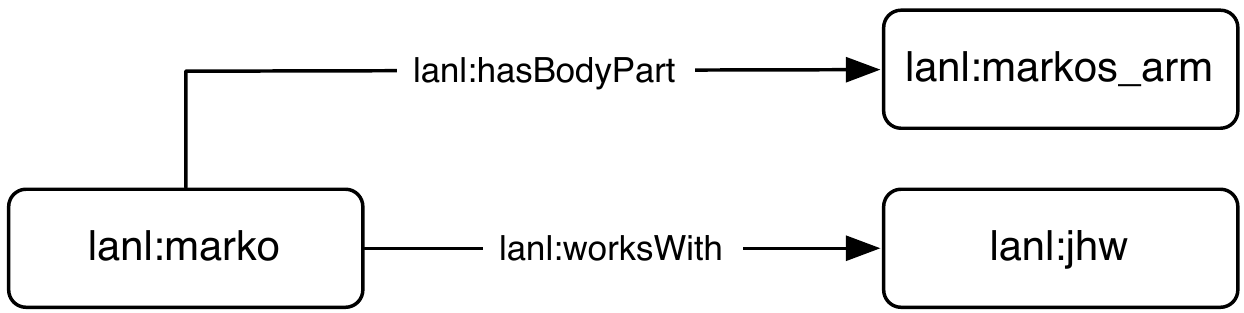}
	\caption{\label{fig:triple}Two RDF triples as an RDF network.}
	\end{center}
\end{figure}

To ease readability and creation, schema and hierarchies are usually prefixed (i.e.~abbreviated). For example, in the following two triples, \ttt{lanl} is the prefix for \ttt{http://www.lanl.gov/people\#}:
\begin{center}
\begin{verbatim}
<lanl:marko, lanl:worksWith, lanl:jhw>
<lanl:marko, lanl:hasBodyPart, lanl:markos_arm>
\end{verbatim}
\end{center}
These triples are diagrammed in Figure \ref{fig:triple}. The union of all RDF triples is the Semantic Web.

The benefit of RDF, and perhaps what is not generally appreciated, is that with RDF it is possible to represent anything in relation to anything by any type of qualified relationship. In many cases, this generality can lead to an uncontrolled soup of relationships; however, thanks to ontology languages such as RDFS and OWL, it is possible to formally constrain the topological features of an RDF network and thus, subsets of the larger Semantic Web.

\section{The RDF Schema and Web Ontology Language\label{sec:onto}}

\begin{figure}[t!]
	\begin{center}
		\includegraphics[width=0.9\textwidth]{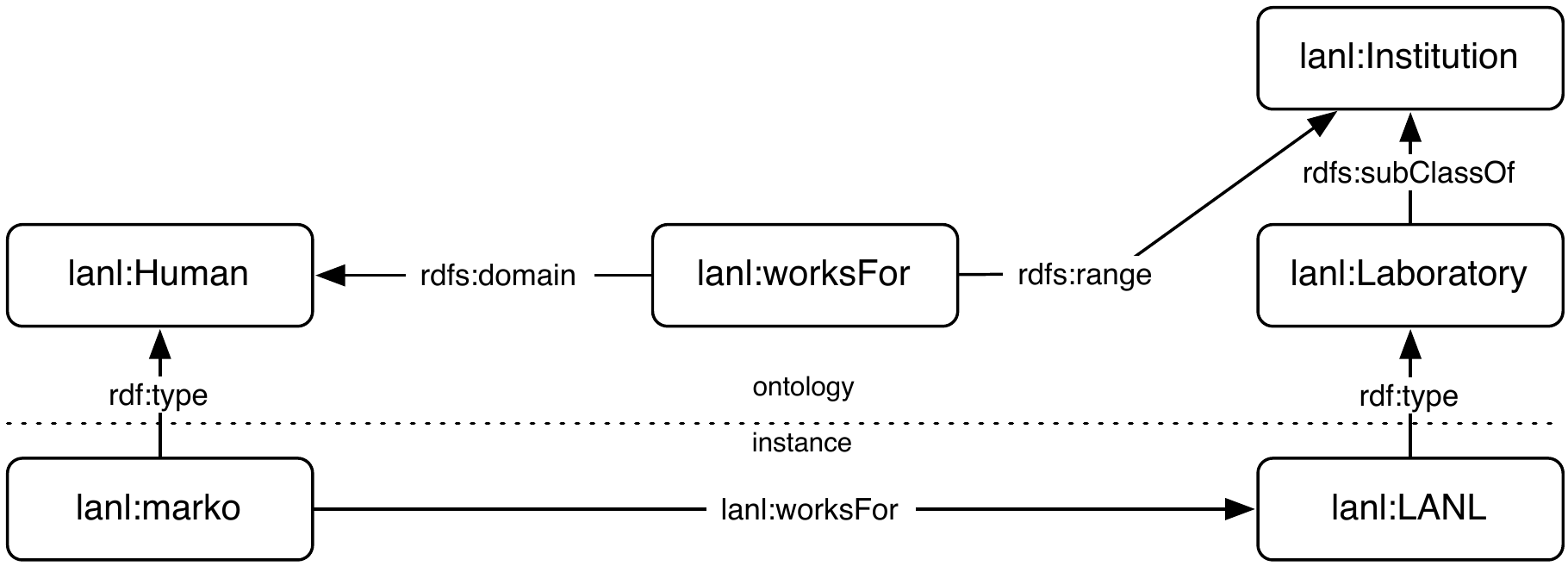}
	\caption{\label{fig:rdfs}The relationship between an instance and its ontology.}
	\end{center}
\end{figure}

The Resource Description Framework and Schema (RDFS) \cite{rdfs:brickley2004} and the Web Ontology Language (OWL) \cite{owlspec:mcguinness2004} are both RDF languages used to abstractly define resources in an RDF network. RDFS is simpler than OWL and is useful for creating class hierarchies and for specifying how instances of those classes can relate to one another. It provides three important constructs: \ttt{rdfs:domain}, \ttt{rdfs:range}, and \ttt{rdfs:subClassOf}\footnote{\ttt{rdfs} is a prefix for \ttt{http://www.w3.org/2000/01/rdf-schema\#}}. While other constructs exist, these three tend to be the most frequently used when developing an RDFS ontology. Figure \ref{fig:rdfs} provides an example of how these constructs are used.
With RDFS (and OWL), there is a sharp distinction between the ontological- and instance-level of an RDF network. The ontological-level defines abstract classes (e.g.~\ttt{lanl:Human}) and how they are related to one another. The instance-level is tied to the ontological-level using the \ttt{rdf:type} predicate\footnote{\ttt{rdf} is a prefix for \ttt{http://www.w3.org/1999/02/22-rdf-syntax-ns\#}}. For example, any \ttt{lanl:Human} can be the \ttt{rdfs:domain} (subject) of a \ttt{lanl:worksFor} triple that has a \ttt{lanl:Institution} as its \ttt{rdfs:range} (object). Note that the \ttt{lanl:Laboratory} is an \ttt{rdfs:subClassOf} a \ttt{lanl:Institution}. According to the property of subsumption in RDFS reasoning, subclasses inherit their parent class restrictions. Thus, \ttt{lanl:marko} can have a \ttt{lanl:worksFor} relationship with \ttt{lanl:LANL}. Note that RDFS is not intended to constrain relationships, but instead to infer new relationships based on restrictions. For instance, if \ttt{lanl:marko} \ttt{lanl:worksFor} some other organization denoted $X$, it is inferred that that $X$ is an \ttt{rdf:type} of \ttt{lanl:Institution}. While this is not intuitive for those familiar with constraint-based database schemas, such inferencing of new relationships is the norm in the RDFS and OWL world.

Beyond the previously presented RDFS constructs, OWL has one primary construct that is used repeatedly: \ttt{owl:Restriction}\footnote{\ttt{owl} is a prefix for \ttt{http://www.w3.org/2002/07/owl\#}}. Example \ttt{owl:Restriction}s include, but are note limited to, \ttt{owl:maxCardinality}, \ttt{owl:minCardinality}, \ttt{owl:cardinality}, \ttt{owl:hasValue}, etc. With OWL, it is possible to state that a \ttt{lanl:Human} can work for no more than $1$ \ttt{lanl:Institution}. In such cases, the \ttt{owl:maxCardinality} restriction would be specified on the \ttt{lanl:worksFor} predicate. If there exist the triples 
\begin{verbatim}
<lanl:marko, lanl:worksFor, lanl:LANL>
<lanl:marko, lanl:worksFor, lanl:LosAlamos>,
\end{verbatim}
an OWL reasoner would assume that \ttt{lanl:LANL} and \ttt{lanl:LosAlamos} are the same entity. This reasoning is due to the cardinality restriction on the \ttt{lanl:worksFor} predicate.

There are two popular tools for creating RDFS and OWL ontologies: Prot\'{e}g\'{e}\footnote{Prot\'{e}g\'{e} available at: http://protege.stanford.edu/} (open source) and Top Braid Composer\footnote{Top Braid Composer available at: http://www.topbraidcomposer.com/} (proprietary).

\section{The Triple-Store\label{sec:store}}

There are many ways in which RDF networks are stored and distributed. In the simple situation, an RDF network is encoded in one of the many RDF syntaxes and made available through a web server (i.e.~as a web document). In other situations, where RDF networks are large, a triple-store is used. A triple-store is to an RDF network what a relational database is to a data table. Other names for triple-stores include semantic repository, RDF store, graph store, RDF database. There are many different propriety and open-source triple-store providers. The most popular proprietary solutions include AllegroGraph\footnote{AllegroGraph available at: http://www.franz.com/products/allegrograph/}, Oracle RDF Spatial\footnote{Oracle RDF Spatial available at: http://www.oracle.com/technology/tech/semantic\_technologies/} and the OWLIM semantic repository\footnote{OWLIM available at: http://www.ontotext.com/owlim/}. The most popular open-source solution is Open Sesame\footnote{Open Sesame available at: http://www.openrdf.org/}.

The primary interface to a triple-store is SPARQL \cite{sparql:prud2004}. SPARQL is analogous to the relational database query language SQL. However, SPARQL is perhaps more similar to the query model employed by logic languages such as Prolog. The example query
\begin{footnotesize}
\begin{verbatim}
SELECT ?x
  WHERE { ?x <lanl:worksWith> <lanl:jhw> . }
\end{verbatim}
\end{footnotesize}
returns all resources that work with \ttt{lanl:jhw}. The variable \ttt{?x} is a binding variable that must hold true for the duration for the query. A more complicated example is
\begin{footnotesize}
\begin{verbatim}
SELECT ?x ?y
  WHERE { 
    ?x <lanl:worksWith> ?y .
    ?x <rdf:type> <lanl:Human> .
    ?y <rdf:type> <lanl:Human> .
    ?y <lanl:worksFor> <lanl:LANL> .
    ?x <lanl:worksFor> <necsi:NECSI> . }
\end{verbatim}
\end{footnotesize}
The above query returns all collaborators such that one collaborator works for the Los Alamos National Laboratory 
(LANL) and the other collaborator works for the New England Complex Systems Institute (NECSI). An example return would be
\begin{footnotesize}
\begin{verbatim}
-------------------------------
|     ?x       |      ?y      |
-------------------------------
| lanl:marko   | necsi:carlos |
| lanl:jhw     | necsi:carlos |
| lanl:jbollen | necsi:carlos |
-------------------------------
\end{verbatim}
\end{footnotesize}

The previous query would require a complex joining of tables in the relational database model to yield the same information. Unlike the relational database index, the triple-store index is optimized for such semantic network queries (i.e.~multi-relational queries). The triple-store a useful tool for storing, querying, and manipulating an RDF network.

\section{A Semantic Network Programming Language and an RDF Virtual Machine\label{sec:comp}}

Neno/Fhat is a semantic network programming language and RDF virtual machine (RVM) specification \cite{rodriguez:gpsemnet2007}. Neno is an object-oriented language similar to C++ and Java. However, instead of Neno code compiling down to machine code or Java byte-code, Neno compiles to Fhat triple-code. An example Neno class is
\begin{footnotesize}
\begin{verbatim}
owl:Thing lanl:Human {
  lanl:Institution lanl:worksFor[0..1];
  
  xsd:nil lanl:quit(lanl:Institution x) {
    this.worksFor =- x;
  }
}
\end{verbatim}
\end{footnotesize}
The above code defines the class \ttt{lanl:Human}. Any instance of \ttt{lanl:Human} can have either $0$ or $1$ \ttt{lanl:worksFor} relationships (i.e.~\ttt{owl:maxCardinality} of $1$). Furthermore, when the method \ttt{lanl:quit} is executed, it will destroy any \ttt{lanl:worksFor} triple from that \ttt{lanl:Human} instance to the provided \ttt{lanl:Institution} \ttt{x}.

\begin{figure}[h!]
	\begin{center}
		\includegraphics[width=0.3\textwidth]{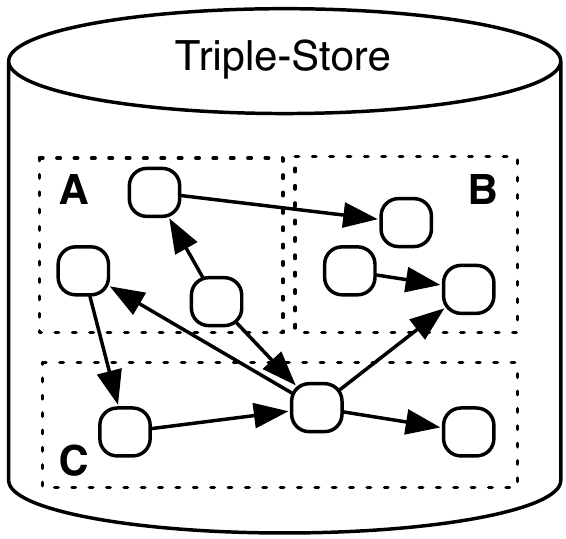}
	\caption{\label{fig:fhat}The Fhat RVM and Neno triple-code commingle with other RDF data.}
	\end{center}
\end{figure}

Fhat is a virtual machine encoded in an RDF network and processes Fhat triple-code. This means that a Fhat's program counter, operand stack, variable frames, etc., are RDF sub-netwoks. Figure \ref{fig:fhat} denotes a Fhat processor (\textbf{A}) processing Neno triple-code (\textbf{B}) and other RDF data (\textbf{C}).

With Neno it is possible to represent both the system model and its algorithmic processes in a single RDF network. Furthermore with Fhat, it is possible to include the virtual machine that executes those algorithms in the same substrate. Given that the Semantic Web is a distributed data structure, where sub-networks of the larger Semantic Web RDF network exist in different triple-stores or RDF documents around the world, it is possible to leverage Neno/Fhat to allow for distributed computing across these various data sets. If a particular model exists at domain $X$ and a researcher located at domain $Y$ needs to utilize that model for a computation, it is not necessary for the researcher at domain $Y$ to download the data set from $X$. Instead, a Fhat processor and associated Neno code can move to domain $X$ to utilize the data and return with results. In Neno/Fhat, the data doesn't move to the process, the process moves to the data.

\section{Conclusion}

This article presented a review of the standards and technologies associated with the Semantic Web that can be used for complex systems modeling. The World Wide Web provides a common, standardized substrate whereby researchers can easily publish and distribute documents (e.g.~web pages, scholarly articles, etc.). Now with the Semantic Web, researchers can easily publish and distribute models and processes (e.g.~data sets, algorithms, computing machines, etc.).

\end{document}